\pdfoutput=1

\documentclass[11pt]{article}

\usepackage{ACL2023}

\usepackage{times}
\usepackage{latexsym}

\usepackage[T1]{fontenc}

\usepackage[utf8]{inputenc}

\usepackage{microtype}

\usepackage{inconsolata}

\usepackage{soul}
\usepackage{tikz}
\usetikzlibrary{shapes.geometric}
\usepackage{framed}
\usepackage{enumitem}
\newlist{RQ}{enumerate}{1}
\setlist[RQ]{label=\textbf{RQ\,\arabic*},ref={RQ\,\arabic*}}
\usepackage{booktabs}
\usepackage{comment}
\usepackage{natbib}
\usepackage{multibib}
\makeatletter

\usepackage{graphicx}
\usepackage{tabularx}
\usepackage{soul}

\usepackage{pifont}
\usepackage{tikz}
\setlist{leftmargin=1mm}
\usetikzlibrary{shapes.geometric, arrows}
\usetikzlibrary{decorations.markings}

\usepackage{xcolor}
\usepackage{hyperref}
 \definecolor{darkblue}{rgb}{0, 0, 0.5}
  \hypersetup{colorlinks=true, citecolor=darkblue, linkcolor=darkblue, urlcolor=darkblue}

\usepackage{xstring}

\usepackage{color}
\usepackage[most]{tcolorbox}
\tcbuselibrary{skins}

\usepackage[export]{adjustbox} 

\usepackage{setspace}
\usepackage[capitalise]{cleveref}

\newcommand*{\Scale}[2][4]{\scalebox{#1}{$#2$}}%

\makeatletter
\newcommand{\DrawLine}{%
  \begin{tikzpicture}
  \path[use as bounding box] (0,0) -- (\linewidth,0);
  \draw[color=blue!75!black,dashed,dash phase=.5pt]
        (0-\kvtcb@leftlower-\kvtcb@boxsep,0)--
        (\linewidth+\kvtcb@rightlower+\kvtcb@boxsep,0);
  \end{tikzpicture}%
  }
\makeatother

\newcommand*{\affaddr}[1]{#1}
\newcommand*{\affmark}[1][*]{\textsuperscript{#1}}
\newcommand*{\email}[1]{\texttt{#1}}

\author{
Vipula Rawte\affmark[1]\thanks{corresponding author}, Prachi Priya\affmark[2], S.M Towhidul Islam Tonmoy\affmark[3], S M Mehedi Zaman\affmark[3], \\ \bf Amit Sheth\affmark[1], Amitava Das\affmark[1]  \\
\affaddr{\affmark[1]AI Institute, University of South Carolina, USA}\\
\affaddr{\affmark[2]Indian Institute of Technology, Kharagpur}\\
\affaddr{\affmark[3]Islamic University of Technology}\\
\email{\{vrawte\}@mailbox.sc.edu}
}

\title{Exploring the Relationship between LLM Hallucinations and Prompt Linguistic Nuances: Readability, Formality, and Concreteness}

\begin{document}
\maketitle
\begin{abstract}
As Large Language Models (LLMs) have advanced, they have brought forth new challenges, with one of the prominent issues being LLM hallucination. While various mitigation techniques are emerging to address hallucination, it is equally crucial to delve into its underlying causes. Consequently, in this preliminary exploratory investigation, we examine how linguistic factors in prompts, specifically readability, formality, and concreteness, influence the occurrence of hallucinations. Our experimental results suggest that prompts characterized by greater formality and concreteness tend to result in reduced hallucination. However, the outcomes pertaining to readability are somewhat inconclusive, showing a mixed pattern.
\end{abstract}

\section{Hallucination in LLMs: An introduction }

The remarkable advantages offered by extensive generative AI models like GPT-4 \citep{brown2020language,openai2023gpt4}, Stable Diffusion \cite{rombach2022high}, DALL-E \cite{ramesh2021zero,ramesh2022hierarchical}, and Midjourney \cite{midjourney} are accompanied by a significant potential for misuse.

The recent and rapid developments in the field of LLMs have been gaining significant attention and use in various applications. This included natural language understanding and generation for chatbots, content generation, translation, summarization, and more. They were also being applied in fields like healthcare, finance, and education. Nevertheless, these LLMs encounter significant hurdles, with one prominent issue being termed \textit{hallucination}. This term describes a situation in which the LLM generates responses that contain factual inaccuracies or fabrications.

\begin{figure}[!ht]
    
\begin{tcolorbox}[enhanced,attach boxed title to top left={yshift=-1mm,yshifttext=-1mm,xshift=8pt},
left=1pt,right=1pt,top=1pt,bottom=1pt,colback=red!5!white,colframe=blue!75!black,colbacktitle=violet!80!black,
  title=Original Prompt,fonttitle=\ttfamily\bfseries\scshape\fontsize{9}{9.6}\selectfont,
  boxed title style={size=small,colframe=violet!50!black},width=0.43\textwidth ]

  \begin{spacing}{0.75}
  \footnotesize Who owned the ships in the Boston Tea Party? 
  \vspace{-3mm}
  \end{spacing}
 
\end{tcolorbox}

\vspace{-3mm}

\begin{tcolorbox}[enhanced,left=1pt,right=1pt,top=1pt,bottom=1pt,  colback=green!5!white,colframe=blue!75!black,colbacktitle=violet!80!black,
  boxed title style={size=small,colframe=violet!50!black}, enlarge left by=0.5cm, width=0.43\textwidth ]

  \begin{spacing}{0.75}
  \footnotesize \textcolor{red}{The ships involved in the Boston Tea Party were British-owned. The three ships—Dartmouth}, Eleanor, and Beaver—that carried the tea cargo were merchant vessels operated by British companies. They were docked in Boston Harbor and were targeted by American colonists who protested against the British-imposed Tea Act by boarding the ships and dumping the tea into the harbor. 
  \vspace{-3mm}
  \end{spacing}
 
\end{tcolorbox}

\vspace{-2mm}

\begin{tcolorbox}[enhanced,attach boxed title to top left={yshift=-1mm,yshifttext=-1mm,xshift=8pt},
left=1pt,right=1pt,top=1pt,bottom=1pt,colback=red!5!white,colframe=blue!75!black,colbacktitle=violet!80!black,
  title=Reformulated Prompt,fonttitle=\ttfamily\bfseries\scshape\fontsize{9}{9.6}\selectfont,
  boxed title style={size=small,colframe=violet!50!black},width=0.43\textwidth ]

  \begin{spacing}{0.75}
  \footnotesize Which \textcolor{blue}{American merchant} owned the ship \textcolor{blue}{Dartmouth} in the Boston Tea Party? 
  \vspace{-3mm}
  \end{spacing}
 
\end{tcolorbox}

\vspace{-3mm}

\begin{tcolorbox}[enhanced,left=1pt,right=1pt,top=1pt,bottom=1pt,  colback=green!5!white,colframe=blue!75!black,colbacktitle=violet!80!black,
  boxed title style={size=small,colframe=violet!50!black}, enlarge left by=0.5cm, width=0.43\textwidth ]

  \begin{spacing}{0.75}
  \footnotesize The ship Dartmouth, one of the vessels involved in the Boston Tea Party, was owned by a partnership of American merchants. \textcolor{teal}{The principal owner of the Dartmouth was Francis Rotch, a wealthy American merchant} and prominent figure in the New England whaling industry. Joseph Rotch, his brother, and William Rotch, their cousin, were also involved in the ownership of the Dartmouth. These merchants were associated with the tea trade and had interests in shipping and commerce. Their ship, the Dartmouth, carried a shipment of British tea to Boston Harbor, where it was targeted by American colonists during the Boston Tea Party protest against British taxation.
  \vspace{-5mm}
  \end{spacing}
 
\end{tcolorbox}
\caption{An illustration of how a ``reformulated prompt'' can aid in addressing the hallucination issue by providing pertinent context. Here, the hallucinated text is highlighted in \textcolor{red}{red}. By introducing additional context highlighted in \textcolor{blue}{blue}, such as ``who'' and ``what'', we modify the prompt to be more formal and concrete. Thus, the newly generated response now incorporates the factually correct (dehallucinated) text, highlighted in \textcolor{teal}{green}.}
\label{fig:first-ex}
\end{figure}

Several mitigation techniques have emerged to address and reduce the occurrence of hallucinations. These techniques can be broadly categorized into two groups: i) Black-box \cite{mündler2023selfcontradictory}, which operates without depending on external grounded knowledge, and ii) Gray-box \cite{zhang2023mitigating,peng2023check,li2023chain}, which incorporates external knowledge to a certain extent.

\ul{Prompt engineering} can play a crucial role in mitigating hallucinations in generative AI models. By providing clear and specific prompts, users can steer the AI model toward generating content that aligns with their intended context or requirements. This can reduce the chances of the model producing hallucinated or inaccurate information.  Prompts can include contextual cues that help the AI model understand the context of the request. This additional context can guide the model in generating responses that are more contextually accurate and less prone to hallucination.  Complex prompts can be used to guide the model through a series of steps, ensuring that it follows a logical sequence of thought and produces coherent responses.

\begin{figure}[htbp]
    \centering
\includegraphics[width=7.7cm,height=7cm]{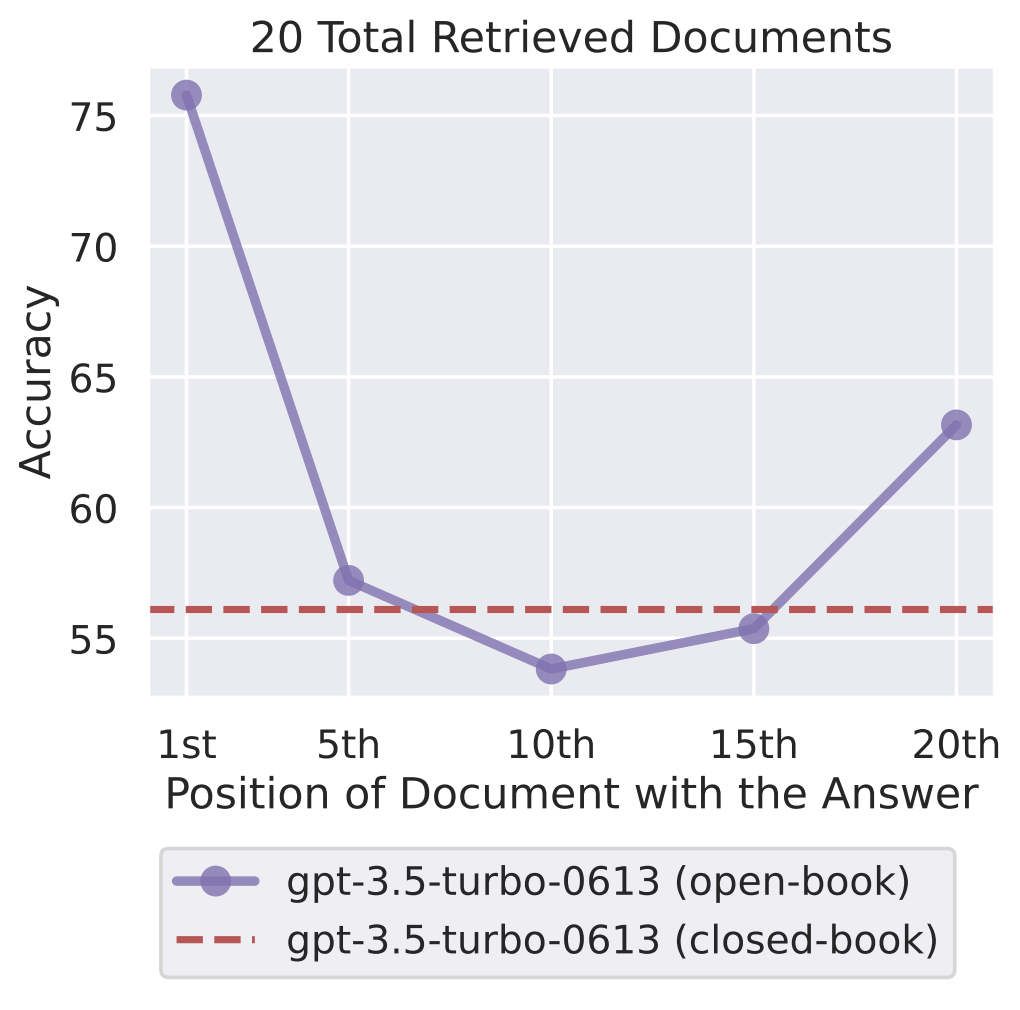}
\vspace{-2mm}
\caption{Empirical results in \cite{liu2023lost} show that the models tend to excel at utilizing pertinent information found at the very start or end of their input context, but their performance notably declines when they need to access and utilize information situated in the middle of their input context.}
    \label{fig:lost}
\end{figure}

The state-of-the-art LLMs have the capability to process lengthy prompts as input. However, findings in \cite{liu2023lost} indicate (see \cref{fig:lost}) that these models tend to perform best when pertinent information is located at the beginning or end of the input context. their performance significantly diminishes when they need to access relevant information in the middle of lengthy contexts. Moreover, as the input context becomes more extended, even models explicitly designed for longer contexts experience a substantial decrease in performance.

In this paper, our primary objective is to explore the impact of the key linguistic attributes of prompts on hallucinations generated in LLMs. The contributions are
as follows: 1) We delineate the broad categories of hallucinations observed in LLMs, as discussed in \cref{sec:hal-cat}. 2) We construct and provide annotations for our dataset, which is derived from tweets related to New York Times events, as detailed in \cref{sec:dataset}. 3) We analyze the relationship between the primary linguistic aspects of prompts, such as their readability, formality, and concreteness, and the occurrence of hallucinations in LLMs, as discussed in \cref{sec:linguistic}.

\section{Types of Hallucination}  \label{sec:hal-cat}

In this study, we explore the following \textit{four} different categories of hallucination. Additionally, we offer examples for each case in which the hallucinated text is marked in \textcolor{red}{red}.

\paragraph{1. Person (P):} The issue of generating fictional characters is discussed in \cite{ladhak2023pre} and \cref{tab:hal-cat-1}.
  
\paragraph{2. Location (L):} The case of generating fictional places is addressed in \cite{ladhak2023pre} and \cref{tab:hal-cat-1}.

\begin{table}[!ht]
\centering
\scriptsize
\begin{tabular}{p{1.1cm}|p{5.6cm}}
\toprule
\textbf{Original}   & Antoine Richard is a former athlete from France who mainly competed in the 100 metres. He was French 100 metre champion on 5 occasions, and also 200 metre winner in 1985. He also won the French 60 metres title 5 times as well.                       \\ \midrule
\textbf{AI-generated} & Athlete Naoki Tsukahara was born in \textcolor{red}{Tokyo, Japan to a Japanese father and French mother}.                         \\ \bottomrule

\end{tabular}
\caption{An example showing how imaginary places such as Tokyo and persons such as father and mother are hallucinated \cite{ladhak2023pre}.}
\label{tab:hal-cat-1}
\end{table}
  
\paragraph{3. Number (N):} Similarly, \cite{varshney2023stitch} delves into the generation of imaginary numbers, as shown in \cref{tab:hal-cat-3}.

\begin{table}[!ht]
\centering
\scriptsize
\begin{tabular}{p{1.1cm}|p{5.6cm}}
\toprule
\textbf{Original}   & Freddie Frith.                      \\ \midrule
\textbf{AI-generated} & He was born in London in \textcolor{red}{1929} and began his racing career in \textcolor{red}{1951}.                       \\ \midrule
\textbf{Fact} & He was born in Grimsby in 1909 and began his career in 1930.                     \\ \bottomrule

\end{tabular}
\caption{Both years 1929 and 1951 are hallucinated.}
\label{tab:hal-cat-3}
\end{table}

\paragraph{4. Acronym (A):}  Additionally, we investigate the potential role of acronyms in prompting the generation of inaccurate responses as illustrated in \cref{tab:hal-cat-4}. 

\begin{table}[!ht]
\centering
\scriptsize
\begin{tabular}{p{1.1cm}|p{5.3cm}}
\toprule
\textbf{Original}   & Robert used PDF for his math homework. \\ \midrule
\textbf{AI-generated} & In the context you provided, ``PDF'' most likely refers to \textcolor{red}{``Portable Document Format,''} which is a common file format used for documents, including math homework assignments. PDF files are widely used for sharing documents because they preserve the formatting and layout across different devices and operating systems. Robert likely used a PDF file for his math homework to ensure that it could be easily shared and viewed by others without compatibility issues.                        \\ \bottomrule
\end{tabular}
\caption{In the mathematical context, ``PDF'' stands for ``Probability Density Function.''.}
\label{tab:hal-cat-4}
\end{table}

\section{Dataset and Annotation} \label{sec:dataset}
To conduct our empirical analysis, where we examine how linguistic properties affect hallucination, we create and annotate a hallucination dataset using the NYT tweets detailed in the following sections.

\subsection{New York Times News Tweets}

We utilize a news dataset, specifically the New York Times (NYT) news events tweets \cite{nyt}. We selected a total of 2,500 tweets. These news tweets serve as our source of factually accurate prompts, which are then presented to the fifteen Large Language Models (LLMs) described in  \cref{sec:llm}.

\subsection{Selection of LLMs} \label{sec:llm}

We have selected 15 contemporary LLMs that have consistently demonstrated outstanding performance across a wide spectrum of NLP tasks. These models include: (i) GPT-4 \cite{openai2023gpt4}
(ii) GPT-3.5 \cite{ChatGPT}
(iii) GPT-3 \cite{brown2020language}
(iv) GPT-2 \cite{radford2019language}
(v) MPT \cite{wang2023multitask}
(vi) OPT \cite{zhang2022opt}
(vii) LLaMA \cite{touvron2023llama}
(viii) BLOOM \cite{scao2022bloom}
(ix) Alpaca \cite{alpaca}
(x) Vicuna \cite{vicuna2023}
(xi) Dolly \cite{dolly}
(xii) StableLM \cite{liu2023your}
(xiii) XLNet \cite{yang2019xlnet}
(xiv) T5 \cite{raffel2020exploring}
(xv) T0 \cite{DBLP:conf/iclr/DeleuKFKBLB22}.

\subsection{Annotation guidelines}

For the purpose of annotating the 2,500 text snippets, we leveraged the services of Amazon Mechanical Turk (AMT) \cite{AMT}. Through this platform, we obtained annotations at the sentence level to identify the different \textit{four} categories of hallucination.

\subsection{Dataset statistics}

Following the annotation process, our dataset statistics for the hallucination categories are presented in \cref{tab:data-stats}.

\begin{table}[!ht]
\centering
\resizebox{0.8\columnwidth}{!}{%
\begin{tabular}{|l|l|}
\hline
\textbf{Category} & \textbf{Hallucinated sentences} \\ \hline
\textbf{Person}   & 14850                         \\ \hline
\textbf{Location} & 13050                          \\ \hline
\textbf{Number}   & 7275                         \\ \hline
\textbf{Acronym}  & 1225                         \\ \hline \hline
\textbf{Total}    & 36910                         \\ \hline
\end{tabular}
}
\caption{Hallucination dataset statistics}
\label{tab:data-stats}
\end{table}

\section{Linguistic Properties of the prompt} \label{sec:linguistic}

Linguistic properties refer to the various characteristics and attributes of language and its components. These properties encompass a wide range of aspects that help define and understand a language.  Some fundamental linguistic properties include: syntactic, semantic, pragmatic, and lexical. Considering these characteristics, we will delve more deeply into the three primary linguistic subtleties in the forthcoming \cref{sec:readability,sec:formality,sec:concrete}.

\subsection{Readability} \label{sec:readability}
\textit{Readability} quantifies the ease with which a text can be comprehended. Several factors, including the text's complexity, familiarity, legibility, and typography, collectively contribute to its readability. 

The Flesch Reading Ease Score (FRES) \cite{flesch1948new} (see \cref{eqn:readability}) is a measure of the readability of a text. It was developed to assess how easy or difficult a piece of text is to read and understand. The score is calculated based on two factors: (a) Sentence Length and (b) Word Complexity.

As shown in the following example, in the first sentence, the language is straightforward, and the sentence is easy to understand, resulting in a high readability score. In contrast, the second sentence contains complex vocabulary and lengthy phrasing, making it more challenging to comprehend, resulting in a lower readability score.

\begin{framed}
\textbf{Easy Readability (High Flesch Reading Ease Score)} \\
\ul{\textbf{Sentence:}} The sun rises in the east every morning.
\end{framed}

\begin{framed}
\textbf{Difficult Readability (Low Flesch Reading Ease Score)}  \\
\ul{\textbf{Sentence:}} The intricacies of quantum mechanics, as expounded upon by renowned physicists, continue to baffle even the most astute scholars.
\end{framed}

\begin{equation} \label{eqn:readability}
\Scale[0.8]{\text{FRES} = 206.835-1.015\left(\frac{\text { total words }}{\text { total sentences }}\right)-84.6\left(\frac{\text { total syllables }}{\text { total words }}\right)}
\end{equation}


To investigate the impact of the readability of the prompt, we pose the following research questions:
\begin{RQ}[align=parleft, leftmargin=!,itemsep=0pt,labelsep=14pt] 
        \item How does the complexity of a prompt's language or vocabulary affect the likelihood of hallucination in LLM-generated responses? \label{RQA}
        \item Does the length of a prompt impact the potential for hallucination, and how does the readability of a long versus a short prompt affect LLM behavior?
        \item How do different LLM architectures (e.g., GPT-3, GPT-4, etc.) respond to prompts of varying linguistic readability, and do they exhibit differences in hallucination tendencies?
\end{RQ}

\subsection{Formality} \label{sec:formality}

The \textit{formality} of language refers to the degree of sophistication, decorum, or politeness conveyed by the choice of words, sentence structure, and overall tone in communication. It is a way to indicate the level of etiquette, respect, or professionalism in a given context. 

In the example given below, both sentences convey an identical message, yet the initial one carries significantly more formality. Such stylistic distinctions frequently exert a more significant influence on the reader's comprehension of the sentence than the literal meaning itself \cite{hovy1987generating}.

\begin{framed}
\textbf{Example of \ul{\textit{formality}} in sentences} \cite{pavlick-tetreault-2016-empirical}
\begin{itemize}

    \item Those recommendations were unsolicited
and undesirable.

    \item that’s the stupidest suggestion EVER.
\end{itemize}
\end{framed}

\textit{Formality} (defined in \cite{heylighen1999formality}) is calculated as given in \cref{eqn:formality}:

\begin{equation} \label{eqn:formality}
    \begin{aligned}
        \text{F} = (\text{noun frequency} + \text{adjective freq.} \\
        + \text{preposition freq.} + \text{article freq.} \\
        \text{- pronoun freq.} \text{ - verb freq.} \\
        \text{- adverb freq.} \text{ - interjection freq.} + 100)/2 \\      
    \end{aligned}
\end{equation}

To examine how the formality of the prompt influences the outcome, we ask the following research inquiries.
\begin{RQ}[align=parleft, leftmargin=!,itemsep=0pt,labelsep=14pt] 
        \item How does the level of formality in prompts influence the likelihood of hallucination in responses generated by LLMs? \label{RQA}
        \item Are there specific categories of hallucination that are more prevalent in responses prompted with formal versus informal language?
        
\end{RQ}

\subsection{Concreteness} \label{sec:concrete}

\textit{Concreteness} assesses the extent to which a word represents a tangible or perceptible concept. As per the theory in \cite{paivio2013dual}, it is suggested that concrete words are easier to process compared to abstract words. The degree of concreteness associated with each word is expressed using a 5-point rating scale that ranges from \ul{abstract} to \ul{concrete}.

A concrete word receives a higher rating and pertains to something that physically exists in reality, i.e. one can directly experience it through senses (smell, taste, touch, hear, see) and actions. An abstract word receives a lower rating and refers to something that isn't directly accessible through your senses or actions. Its meaning is dependent on language and is usually elucidated by employing other words since there's no straightforward method for direct demonstration. 

\begin{framed}
    \textbf{Examples of \ul{\textit{concrete}} words} \\
    Apple, Dog, Chair, Book, Water, Mountain, Car
\end{framed}

\begin{framed}
   \textbf{Examples of \ul{\textit{abstract}} words}  \\
   Justice, Love, Happiness, Courage, Friendship, Wisdom, Equality, Democracy
\end{framed}

Concreteness ratings for 37,058 individual English words and 2,896 two-word expressions  (i.e., a total of 39,954) are provided in \cite{brysbaert2014concreteness}. Since these ratings are at the word level, we compute the concreteness of a sentence by taking an average as described in \cref{eqn:concrete}.

\begin{equation} \label{eqn:concrete}
    \begin{aligned}
        \text{concreteness of a sentence containing $n$ tokens} = \\
    \frac{\sum_{i=1}^{n}\text{concreteness rating}_{i}}{n}
    \end{aligned}
\end{equation}

In order to explore the influence of the prompt's concreteness on the study, we present the following research questions.
\begin{RQ}[align=parleft, leftmargin=!,itemsep=0pt,labelsep=14pt] 
        \item How does the level of linguistic concreteness in a prompt impact the probability of hallucination in LLMs? \label{RQA}
        \item Do LLMs tend to hallucinate less when provided with prompts that include specific details and constraints?
        \item Are LLMs more prone to hallucination when given abstract or vague prompts compared to concrete and specific prompts?
\end{RQ}

\section{Our findings}
To investigate how the linguistic characteristics of prompts affect the generation of hallucinations in LLMs, we initially define the ranges for three specific scores, as outlined in \cref{tab:range}. A comprehensive analysis of these findings is presented in the following sections.

\begin{table}[htbp]
\small
    \centering
    \begin{tabular}{c|c|c|c}  \toprule
    \bf Range $\rightarrow$ \\ \bf Linguistic Aspect $\downarrow$ &  \bf Low &   \bf Mid  &  \bf High\\ \midrule
    \bf Readability   &  0-30 & 31-70 & 71-100  \\ \midrule
    \bf Formality     &  0-30 & 31-70 & 71-100  \\ \midrule
    \bf Concreteness  &  1-2.5 & 2.5-3.5 & 3.5-5  \\ \bottomrule
    \end{tabular}
    \caption{Range(s) for three linguistic aspects of the prompt.}
    \label{tab:range}
\end{table}

\subsection{Effects of \textit{readability} on hallucination in LLMs}

\begin{figure*}[!ht]
    \centering
\includegraphics[width=15cm,height=9.5cm]{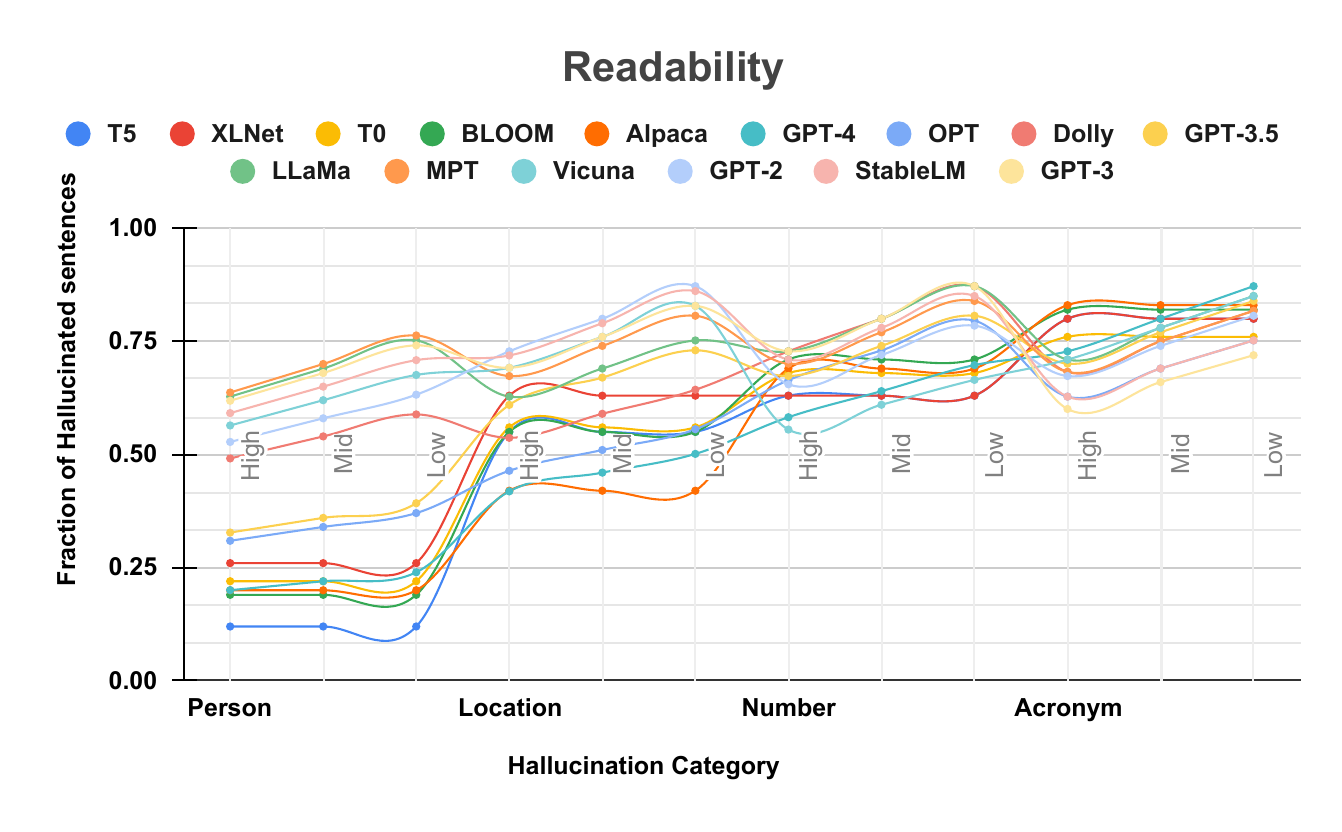}
\vspace{-2mm}
\caption{Hallucination vs Readability}
    \label{fig:readability}
\end{figure*}

The figure (see \cref{fig:readability}) illustrates our empirical findings and the following are the main insights that address the research questions posed earlier in \cref{sec:readability}.

\begin{tcolorbox}
[colback=orange!5!white,colframe=orange!65!black,title=\textbf{\footnotesize \textsc{\ul{Effects of readability on LLM's hallucination}}:}]

\begin{itemize}
[leftmargin=1mm]
\setlength\itemsep{0em}
    \item[\ding{224}] {\footnotesize 
    {\fontfamily{phv}\fontsize{8}{9}\selectfont
   Prompts that are easier to read tend to have fewer instances of hallucinations. }} 
    
    \item[\ding{224}] {\footnotesize 
    {\fontfamily{phv}\fontsize{8}{9}\selectfont
   Some difficult-to-read prompts, but more formal also hallucinate less.}}

     \item[\ding{224}] {\footnotesize 
    {\fontfamily{phv}\fontsize{8}{9}\selectfont
     Hence, the results regarding readability are somewhat uncertain, displaying a combination of findings.
    }}
\end{itemize}
\end{tcolorbox}

\subsection{Effects of \textit{formality} on hallucination in LLMs}

\begin{figure*}[!ht]
    \centering
\includegraphics[width=15cm,height=9cm]{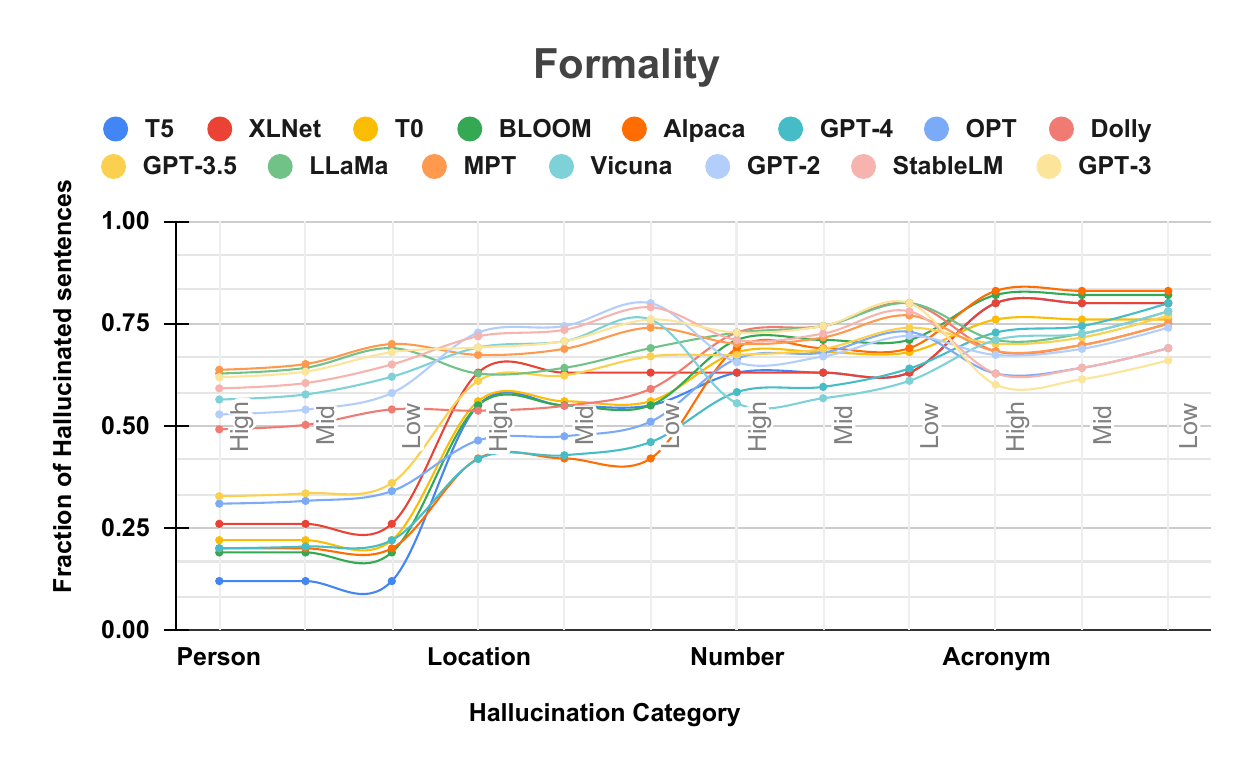}
\vspace{-2mm}
\caption{Hallucination vs Formality}
    \label{fig:formality}
\end{figure*}

\cref{fig:formality} represents our empirical findings.The following points outline the primary insights that respond to the research queries introduced in \cref{sec:formality}.

\begin{tcolorbox}
[colback=orange!5!white,colframe=orange!65!black,title=\textbf{\footnotesize \textsc{\ul{Effects of formality on LLM's hallucination}}:}]

\begin{itemize}
[leftmargin=1mm]
\setlength\itemsep{0em}
    \item[\ding{224}] {\footnotesize 
    {\fontfamily{phv}\fontsize{8}{9}\selectfont
   Formal language prompts typically exhibit a lower propensity for generating hallucinatory content. }} 
    
    \item[\ding{224}] {\footnotesize 
    {\fontfamily{phv}\fontsize{8}{9}\selectfont
    Our findings demonstrate how utilizing more formal prompts can address hallucinations in the \bf{Name} and \textbf{Location} categories.}}

     \item[\ding{224}] {\footnotesize 
    {\fontfamily{phv}\fontsize{8}{9}\selectfont
    The linguistic impacts of the prompts become more evident in LLMs such as GPT-4, OPT, and subsequent versions.
    }}
\end{itemize}
\end{tcolorbox}

\subsection{Effects of \textit{concreteness} on hallucination in LLMs}

\begin{figure*}[!ht]
    \centering
\includegraphics[width=15cm,height=9.5cm]{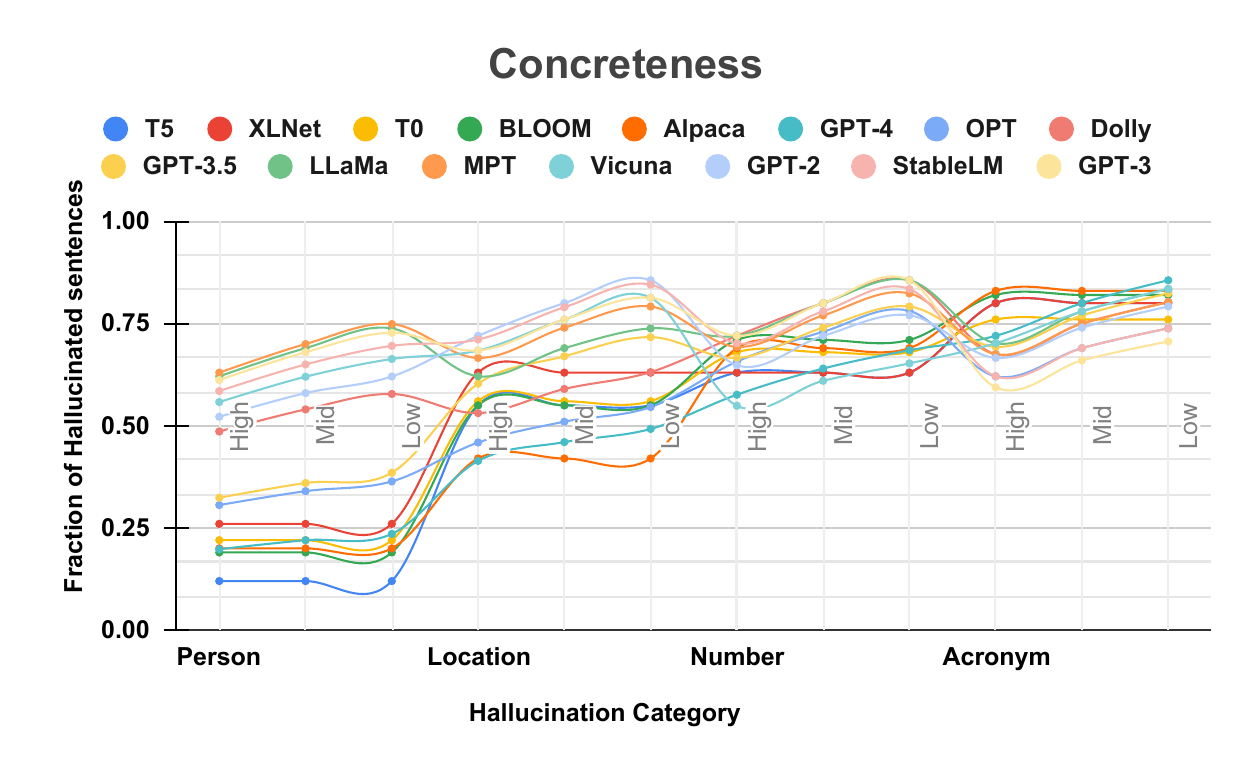}
\vspace{-2mm}
\caption{Hallucination vs Concreteness}
    \label{fig:concrete}
\end{figure*}

\cref{fig:concrete} shows our experimental results. The following section highlights the core insights that address the research inquiries introduced in \cref{sec:concrete}.

\begin{tcolorbox}
[colback=orange!5!white,colframe=orange!65!black,title=\textbf{\footnotesize \textsc{\ul{Effects of concreteness on LLM's hallucination}}:}]

\begin{itemize}
[leftmargin=1mm]
\setlength\itemsep{0em}
    \item[\ding{224}] {\footnotesize 
    {\fontfamily{phv}\fontsize{8}{9}\selectfont
   Prompts that use clearer and more specific language tend to generate fewer hallucinations. }} 
    
    \item[\ding{224}] {\footnotesize 
    {\fontfamily{phv}\fontsize{8}{9}\selectfont
    Our results show that incorporating more specific and concrete terms into the prompts effectively reduces hallucinations in the \bf Number and Acronym categories.}}

     \item[\ding{224}] {\footnotesize 
    {\fontfamily{phv}\fontsize{8}{9}\selectfont
    Just as we observed with our formality findings, the impact of concrete prompts becomes increasingly apparent in advanced LLMs like GPT-4, OPT, and their later iterations.
    }}
\end{itemize}
\end{tcolorbox}

\section{Conclusion}

In this preliminary research study, we begin by categorizing the primary types of hallucinations present in LLMs. Subsequently, we compile our dataset by utilizing New York Times news tweets, aligning with these established categories. Language intricacies assume a crucial role in the comprehension of language. Therefore, we delve into the examination of three significant linguistic dimensions: readability, formality, and concreteness, and their potential influence on the occurrence of hallucinations in LLMs.

\bibliography{custom}
\bibliographystyle{acl_natbib}

\end{document}